\documentclass[aps,floats,twocolumn,epsf,prb,showpacs]{revtex4-2}

\usepackage{latexsym,amsmath,amssymb,bm,graphicx,amsfonts,booktabs}

\usepackage[plain]{algorithm}
\usepackage[noend]{algpseudocode}
\usepackage{bm}

\makeatletter\def\Dated@name{}\makeatother

\newcommand{\norm}[1]{\left\lVert#1\right\rVert}
\def\BState{\State\hskip-\ALG@thistlm}
\DeclareMathOperator{\sign}{sign}
\DeclareMathOperator{\mean}{mean}

\begin{document}

\title{Genetic AI: Evolutionary Games for \emph{ab initio} dynamic Multi-Objective Optimization}
\author{P. Wissgott$^{1}$}
\affiliation{$^1$ danube.ai solutions gmbh, 1040 Vienna, Austria\\
}

\begin{abstract}
We introduce Genetic AI, a novel method for multi-objective optimization without external parameters or predefined weights. The method can be applied to all problems that can be formulated in matrix form and allows for a data-less training of AI models. Without employing predefined rules or training data, Genetic AI first converts the input data into genes and organisms. In a simulation from first principles, these genes and organisms compete for fitness, where their behavior is governed by universal evolutionary strategies. We present four evolutionary strategies: Dominant, Altruistic, Balanced and Selfish and show how a linear combination can be employed in a fully self-consistent evolutionary game. Investigating fitness and evolutionary stable equilibriums, Genetic AI helps solving optimization problems with a set of predefined, discrete solutions that change dynamically. We show the universality of the approach on two decision problems.
\end{abstract}

\maketitle

\section{Introduction}
\label{sec:intro}
In the past two decades, the rise of Big Data has proven pivotal for many industries and markets. With this shift to the realms of data comes a huge need for optimization and data analysis e.g. for automation, pattern recognition, prediction, consumer needs or decision making.

In many of these problems, one finds methods falling into two classes of algorithms: (i) optimization, where one aims to find the optimum in a manifold of solutions~\cite{Yang1970} or (ii) Machine Learning~(ML), where one trains AI models to gain knowledge about a system~\cite{Choi2020}. 

\begin{figure}[tb]
  \centering
  \includegraphics[width=\linewidth]{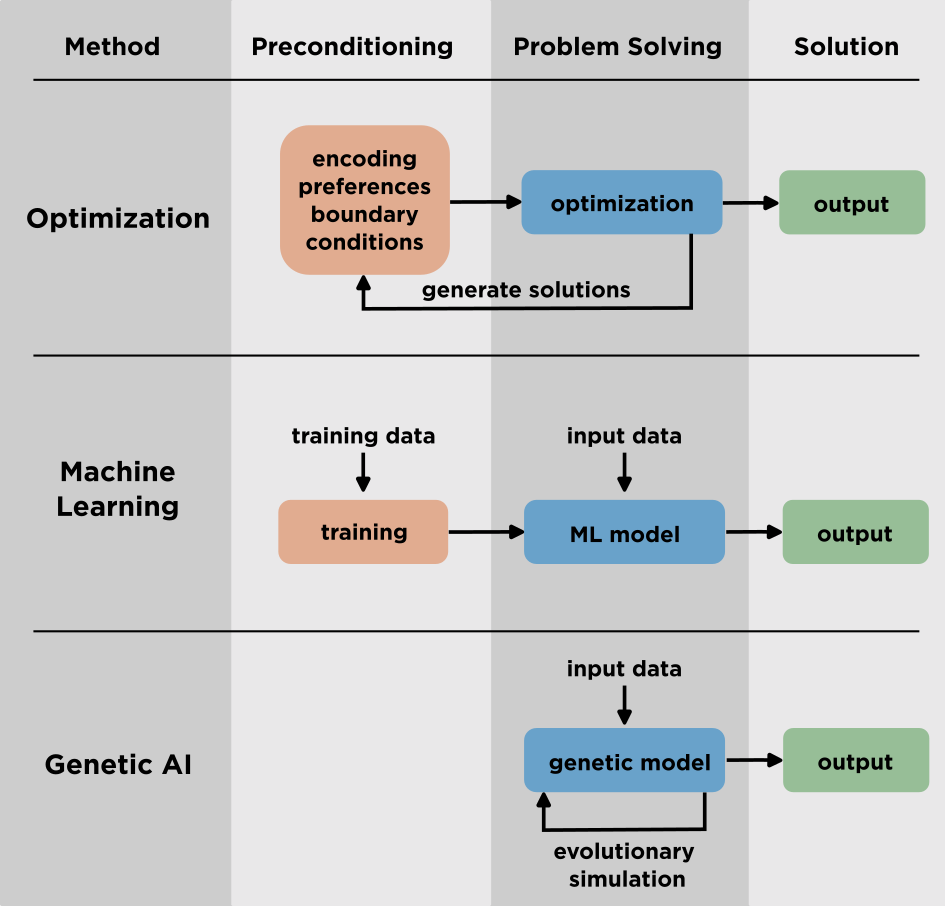}
  \caption{Methodological comparison of the workflow in optimization algorithms, Machine Learning and Genetic AI.}
  \label{fig:comparisonoptMLGAI}
\end{figure}

On the one hand, in many optimization algorithms, there is a setup phase \emph{before} the actual optimization. For example, in evolutionary multi-objective optimization~(EMO), one usually converts the data to an evolutionary picture where the applied encoding usually depends on the problem field~\cite{Deb2001}. Additionally, depending on the algorithmic variation of EMO, one may select rules which objective dominates another or chooses weights in a weighted cost/fitness function~\cite{Friedrich2011}. Fundamentally, in these kind of algorithms, there is a certain form of preconditioning, influencing the dynamics of the optimization. While for some systems, predefining multi-objective behavior may work out well, the same rules may fail to meet quality requirements in other applications. See Fig.~\ref{fig:comparisonoptMLGAI} for a visualization of the workflow in general optimization. Note that one defining feature in many optimization algorithms is the lack of input data, since they generate their own solutions in a feedback loop.

On the other hand, algorithms from the field of Machine Learning~(ML) depend much less on predefined quantities. In a training phase, they learn the 'right' behavior for a certain data problem~\cite{Sarker2022}. After an ML model has been trained for a specific application, it can convert input data to an output through inference~\cite{Yuan2024}. While this approach makes ML usually more universal than rule-based algorithms, choosing certain training data may be interpreted as statistically preconditioning the AI model. Consequently, models in ML highly depend on the choice of training data, biasing the result an ML algorithm will deliver. See Fig.~\ref{fig:comparisonoptMLGAI} for a visualization of the workflow in general ML algorithms. Note that the training phase and the inference phase~(converting input data to output) are distinct steps that are usually covered by rather different algorithmic strategies.

In this paper, we propose a novel approach, apart from rule-based optimization and ML: Genetic AI. Our method identifies a result from input data without any   statistical preconditioning or training. In contrast, Genetic AI solves data problems by an \emph{ab initio} approach:  converting the input data into a universal, evolutionary representation allows to run autonomous simulations without the need of any predefined behavior~\footnote{We are using the expressions 'ab initio' and 'from first principles' in this paper. The argumentation follows theories in solid state physics which aim to compute properties of materials without external parameters~\cite{Hohenberg1964, Kohn1965, Held2001}. Quite similar, in Genetic AI, we aim to gain understanding of data problems without any external parameters.}. See Fig.~\ref{fig:comparisonoptMLGAI} for a comparison of the algorithmic workflow in Genetic AI. Methodologically, Genetic AI somehow 'stands' in between (evolutionary) optimization and ML: using input data is reminiscent of ML, while employing an (evolutionary) feedback loop relates to optimization. Additionally, as we will see below, Genetic AI uses replicator equations and 'game rounds' as introduced in evolutionary game theory~(EGT, \cite{MaynardSmith1973, Hofbauer2011}).

In the current formalism, Genetic AI can be applied to any discrete optimization problem where possible solutions can be represented in matrix form. Through evolutionary dynamics, Genetic AI not only finds the optimal solutions but also tests universal symmetries, uniqueness and relations in the data to gain understanding of them. Instead of predefining algorithmic  behavior, these simulations open up a new route to fundamentally analyze the mechanics of a system described by data.


How do we obtain knowledge of a system without defining the 'right' behavior in terms of predefined weights or training data? We achieve this by 'translating' input data directly into a biological system that self-consistently reaches an evolutionary equilibrium. In this paper, we present our approach as follows: in Sec.~\ref{sec:formalism}, we describe how the input data can be converted into an evolutionary picture of genes and organisms. Then, we proceed by introducing the details of an evolutionary simulation in Sec.~\ref{sec:evolutionarysimulation}. The dynamics of the evolutionary model are governed by evolutionary strategies, described in Sec.~\ref{sec:evolutionarystrategies}. After running the simulation with these strategies, we obtain data features that are deemed more relevant. In Sec.\ref{sec:numericalexperiments}, we discuss the dynamics of two examples. Furthermore, in Sec.~\ref{sec:self-consistentdynamics}, we show a method to self-consistently converge to a stable state without external parameters or preconditions.

Without training data, Genetic AI becomes in a sense a more 'autonomous' AI than general ML. This has also philosophical implications: What is the 'right' behavior if not preconditioned by training? Since this paper has introductory purpose, we will leave a deeper analysis of these implications to future work. Note that one \emph{can} reintroduce a a certain kind of training into Genetic AI~(see Sec.~\ref{sec:mixingtroughtraining}).

It is important to emphasize that Genetic AI goes beyond purely statistical approaches. Instead of statistical behavior, the individual structure of the input data at hand determines the result of the evolutionary simulation. In particular, statistical outliers that would be removed in other methods, often play an important role in Genetic AI. As in real biological systems, a single variation in a gene can effect the whole population~(see Sec.~\ref{sec:realworldexample}).

\section{Background}
\label{sec:background}
Applying evolutionary concepts to non-biological problems has a long history in science and engineering. Most methods, from the early evolutionary algorithms~(EA,\cite{DeJong1997}) to the field of neuroevolution~\cite{Galvan2021}, share an abstract interpretation of data. In this sense, these approaches are rather universal in terms of problems they can be applied to, as long as the model's parameters can somehow be encoded into the computational formulation in a meaningful way. Following this philosophy, Genetic AI aims to 'translate' a data problem into an evolutionary 'game' of genes and organisms.

In addition to evolutionary algorithms, Genetic AI is also related to EMO~\cite{Deb2001}. Quite similar to EMO, one objective of Genetic AI is to sort solutions according to their fitness and taking into account trade-offs between different objectives. There are however major methodological differences between Genetic AI and existing methods in EMO: (i) there is no continuous solution space or generation of new solutions in Genetic AI - only the given, discrete 'solutions' from the input data take part in the evolutionary simulation, (ii) in contrast to weighted approaches in EMO, Genetic AI dynamically adapts the weights themselves during the simulation, (iii) in general Genetic AI, there are no preferences for objectives defined before, during or after the simulation, (vi) in Genetic AI, there are no externally dominated/non-dominated rules but behavioral strategies.



Though we created Genetic AI from scratch without using any scientific references, we employ many terms from EGT~\cite{MaynardSmith1973, MaynardSmith1974, MaynardSmith1981, Axelrod1984, Berger2005, Hofbauer2011} in our formulation. This has two major reasons, (i) we can stick to well-known vocabulary in describing Genetic AI (though most quantities differ at least slightly from their direct analogue in EGT); (ii) the formalisms introduced in EGT are very versatile for general evolutionary simulations.

\section{Formalism}
\label{sec:formalism}
Let us in the following restrict to discrete optimization problems where the solutions can be represented as rows in a $(n\times m)$-matrix form
\begin{align}
\label{eqref:rawinputdata}
X_p = 
\begin{pmatrix} 
a_{11}      &       \cdots      &    a_{1m}     \\    
\vdots      &       \ddots      &    \vdots \\        
a_{n1}      &       \cdots      &    a_{nm}  \\
\end{pmatrix} \text{ with } X_p\in \mathcal{D}
\end{align}
where $a_{ij}$ can be in general any form of structured data element and $\mathcal{D}$ denotes a problem family where all $X_p$ share the same principle structure of columns. 

In some methods for multi-objective optimization~(MOO) one would pre-select weights analysing the problem family $\mathcal{D}$. In contrast, in ML algorithms, one would try to obtain as many data packages $X_p$ as available to train the wanted optimization dynamics according to predefined behavior. In Genetic AI, every $X_p$ poses an independent, discrete optimization problem that is solved by evolutionary mechanisms.
%

\subsection{Genes and Organisms}
\label{sec:genesorganisms}

We now identify genes and organisms in our data. In biological systems, organisms can be understood as replicator machines of genes~\cite{Dawkins2006}. In this interpretation, organisms are \emph{assembled} according to the plan stored in the genes.

Quite similar, solutions or data sets are just a 'wrapper' of a list of logically related data features. Consequently, a solution is 'built up' from data features analogously than organisms are built up from genes. Hence, we consider the mapping
\begin{align}
 \text{gene}&\leftrightarrow \text{data feature}\label{eqn:genetodatafeature}\\
 \text{organism}&\leftrightarrow \text{data set}.\label{eqn:organismtodataset}
\end{align}
At this point it is important to differentiate between a data feature, i.e. a data column in Eq.~\eqref{eqref:rawinputdata}, and a data element $a_{ij}$, constituting the smallest unit of information in the system. 

In the genetic picture, gene variants define how a certain gene is expressed in a specific organism of the population. Taking for example a gene that is related to the size, a specific gene variant makes the particular organism larger or smaller. In the data picture, data elements define how a certain data feature is 'expressed' in a particular data set. Hence, whereas data feature map to genes in Genetic AI, data elements map to gene variants
\begin{align}
 \text{genes variant}&\leftrightarrow \text{data element}.\label{eqn:genevarianttodataelement}
\end{align}
In the following we will switch between evolutionary and data picture smoothly and will use the analogies Eq.\eqref{eqn:genetodatafeature}-\eqref{eqn:genevarianttodataelement} interchangeably.

\subsection{Fitness functions \& Population}
\label{sec:fitnessfunctions}

In Genetic AI, there are three different levels of fitness sorted from lowest to highest organisational hierarchy (i) the gene variant fitness $\phi$, (ii) the gene fitness $\gamma$, (iii) the organism fitness $F$. We will generally assume that all fitness functions are positive functions 
\begin{align}
 f: x \mapsto f(x) \text{ with } 0\leq f(x) \leq 1,\label{eqn:rangefitness}
\end{align}
where $f$ denotes the gene variant fitness function~$\phi$, the gene fitness~$\gamma$ or the organism fitness function Formalism~$F$, respectively. Note that during iterations of the evolutionary simulation, the gene fitness may violate the range Eq.~\eqref{eqn:rangefitness}, but will always be normalized at the end of the replication cycle~(see further below).

In the gene variant fitness $\phi_j(a_{ij})$ the fitness function $\phi_j$ define how 'fit' the data element $a_{ij}$ is compared to the other data elements within the same data feature $y_j$. In the simplest case, a gene variant fitness function is boolean
\begin{align}
\phi_j^{bool}(a_{ij})=\left\{
    \begin{array}{ll}
        0 & \text{if the data feature } j \text{ is not present}\\
        1 & \text{if the data feature } j \text{ is present,}\\
    \end{array}
    \right. \label{eqn:booleangenevariant}
\end{align}
where $i$ again denotes the $i$th data set. 
The choice of the right fitness function $\phi_j$ for a gene variant of course influences the behavior of the system in the evolutionary simulation.

It is important to mention that the gene variant fitness is determined once, \emph{before} the evolutionary simulation. In particular, it does not change during the iterations and is thus applied as a preprocessing step on the input data $X_p$ Eq.~\eqref{eqref:rawinputdata}. In the following, we will use the gene variant fitness matrix
\begin{align}
\Phi(X_p) = 
\begin{pmatrix} 
\phi_1(a_{11})      &       \cdots      &    \phi_m(a_{1m})     \\    
\vdots      &       \ddots      &    \vdots \\        
\phi_1(a_{n1})      &       \cdots      &    \phi_m(a_{nm})  \\
\end{pmatrix}.
\label{eqn:genevariantfitnessmatrix}
\end{align}
We will also denote $\Phi(X_p)$ as the population and use the analogy
\begin{align}
 \text{population}&\leftrightarrow \text{input data with gene variant fitness}.\label{eqn:populationtofitnessrawdata}
\end{align}
Note that in Genetic AI the population does not change in the evolutionary simulation. Keeping the population unchanged throughout the simulation is a different approach compared to EMO~\cite{Deb2001} or EGT~\cite{MaynardSmith1973}.


Analogously to the population, we can now also define the genes~(the genetic representation of the data columns)
\begin{align}
  \bm{g}_j = \left[\phi_j(a_{1j}),\dots, \phi_j(a_{nj})\right] \text{ with } 1\leq j \leq m,
\end{align}
and organisms~(the genetic representation of the data rows or solutions)
\begin{align}
  \bm{\omega}_i = \left[\phi_1(a_{i1}),\dots, \phi_m(a_{im})\right] \text{ with } 1\leq i \leq n,
\end{align}
in terms of the gene variant fitness functions.

The second fitness introduced is the gene fitness. For simplicity, we will use a normalized vector of positive gene fitness $\bm{\gamma}$ with
\begin{align}
  \sum_{j=1}^m \gamma_j = 1 \text{ where } 0\leq \gamma_j \leq 1 \text{ for } 1\leq j \leq m, \label{eqn:definitiongenefitness}
\end{align}
where $\gamma_j$ denotes the gene fitness of the $j$th data feature. The gene fitness values represent central quantities in Genetic AI. As the simulation progresses, some genes will become more dominant or recessive.

In general, the third fitness, the organism fitness function, may depend on several quantities, e.g. the history of gene fitness values. In this paper, we will limit ourselves for simplicity to a linear approach
\begin{align}
  r_i^{(k)} = \bm{\omega}_i\cdot \bm{\gamma}^{(k)} = \sum_{j=1}^m \gamma_j^{(k)} \phi_j(a_{ij}), \label{eqn:linorganismfitness},
\end{align}
where $(k)$ denotes the $k$th iteration of the evolutionary simulation. Hence, the organism fitness is a linear combination of the gene variant fitness and the current gene fitness. In Genetic AI, one main objective usually is to determine a converged set of gene fitness $\bm{\gamma}^{(k)}$ by evolutionary simulation. This means that while $\phi_j(a_{ij})$ stays fixed throughout the iterations, $\bm{\gamma}^{(k)}$ and, with Eq.~\eqref{eqn:linorganismfitness} also $r_i^{(k)}$, evolves during the simulation.


In an evolutionary interpretation this means that the more gene variant fitness an organism for a more valuable gene has, the fitter the organism will be in the population. In the example of size above that would mean that a larger organism would be fitter, depending on how the gene responsible for 'size' is deemed important in the population.

Quite similar to EGT, where system setups may look simple on the outset~\cite{Berger2005}, also the restriction to a linear organism fitness may appear as an oversimplification at the first glance. However, analogously to EGT, the dynamics of all organisms and genes usually leads to a complex evolution of a system even for linear organism fitness as we will see further below\footnote{Note that the approximation of taking a linear organism fitness shows similarities to applying the local density approximation~(LDA) in physical and chemical simulations~\cite{Hohenberg1964, Held2001}). In LDA, one neglects (some) correlations by replacing complex electronic orbitals by a single function, the electronic density. In Genetic AI, by taking a linear fitness, we neglect non-local inter-organism correlations to the organism fitness.}.

At this point, it is useful for the understanding of Genetic AI to compare some basic concepts with EMO~\cite{Friedrich2011}. While in EMO the letter $X$ is often denoting a subset of the solution space $S$, it represents the input data in Genetic AI with $X_p=S$, since there are no other discrete solutions allowed as defined by the input. Furthermore, the concept of the Pareto front as a surface of points, minimizing individual objectives, becomes less useful in Genetic AI. This is because by dynamically adapting the gene fitness $\bm{\gamma}$, we are effectively changing the search space itself. Hence, instead of an optimization with fixed objectives, the evolutionary simulation 'distorts' the solution space until the evolutionary behavior reaches a stable state. In the next section, we introduce the necessary quantities describing this dynamics.

\section{Evolutionary Simulation}
\label{sec:evolutionarysimulation}

In Genetic AI, organisms $\bm{\omega}_i$ and genes $\bm{g}_j$ compete for the available fitness in the system. Starting from an initial gene fitness $\bm{\gamma}^{(k=0)}$ one iteratively obtains new values $\bm{\gamma}^{(k+1)}$. With the new gene fitness values, one can then obtain the organism fitness values via Eq.~\eqref{eqn:linorganismfitness}.

It follows from Eq.~\eqref{eqn:rangefitness} and Eq.~\eqref{eqn:definitiongenefitness} that the total gene fitness cannot be created or destroyed. Hence, if one gene $g_j$ increases in fitness $\gamma_j$, it comes at the cost of some other genes. This follows the argumentation of Dawkins~\cite{Dawkins2006}, but since we are investigating data, it leads to interesting observations. What does it mean that one data feature becomes 'fitter' than the others? In many cases, it means that the fitter feature is more \emph{relevant} wrt. the others for the data analysis at hand.

From a algorithmic point of view, Genetic AI iteratively updates the gene fitness $\bm{\gamma}^{(k)}$ until either convergence or a maximum number of iterations is reached~(see also the pseudo code in Alg.\ref{alg:geneticaisimulation}).
\begin{algorithm}[H]
\caption{Genetic AI simulation}
\begin{algorithmic}[1]
\Function{Simulation}{input data $X$, GS+OS, $\bm{\gamma}^{(0)}$} 
\State $\Phi(X) \gets X$ normalize data via Eq.\eqref{eqn:genevariantfitnessmatrix}
\State $k \gets 0$
\While{$k<maxiteration$} 
\ForAll{genes} 
  \State $\Delta^{g}=$\Call{ Update}{gene, GS}
\EndFor
\ForAll{organisms} 
  \State$\Delta^{\omega}=$\Call{ Update}{organism, OS}
\EndFor
\State $\Delta = \Delta^{g} + \Delta^{\omega}$
\State $\bm{\gamma}^{(k+1)} \gets \Delta,\bm{\gamma}^{(k)}$ via Eq.~\eqref{eqn:replicator1}+\eqref{eqn:replicator2}
\State $\bm{r}^{(k+1)}\gets \bm{\gamma}^{(k+1)}, \Phi(X)$ via Eq.\eqref{eqn:linorganismfitness}
\If{$\|\bm{\gamma}^{(k+1)}-\bm{\gamma}^{(k)}\|<\varepsilon$} 
\State break
\Else 
\State $k\gets k+1$
\EndIf

\EndWhile
\State \Return $\bm{\gamma}^{(k+1)}$, $\bm{r}^{(k+1)}$
\EndFunction
\end{algorithmic}
\label{alg:geneticaisimulation}
\end{algorithm}

It is important to emphasize that Genetic AI is different from optimization algorithms that search for local minima in a multi-dimensional solutions space. This is because in Genetic AI one does not create new data sets in a (bounded) solution space.  Quite contrary, the discrete data or population $\Phi(X_p)$ is fixed throughout the simulation. 

Proceeding in this argumentation, it becomes clear that another input data $X_{p'}$, may lead to other gene fitness values $\bm{\gamma}'$ and hence to other optimal solutions. In this sense, Genetic AI allows for a dynamic optimization depending on fixed packages $X_p$ of data sets. The solutions included in a packages are analyzed wrt. to each other and \emph{not} wrt. other data packages~(like e.g. training data). The mutual independence of the evolutionary simulations that comes with this \emph{ab initio} approach has important consequences when it comes to entities like data bias and opens up new perspectives in optimization and data analysis.

\subsection{Replicator Equations}

In EGT, replicator equations are used to investigate the game dynamics and which how a set of strategies perform in the game~\cite{Hofbauer2011}. In contrast, in Genetic AI, there are two fixed types of strategies: (i) the gene strategy~(GS) and (ii) the organism strategy~(OS). 

Replicator equations prove to be very useful to analyze the dynamics of the gene fitness $\bm{\gamma}^{(k)}\rightarrow \bm{\gamma}^{(k+1)}$. To that end, let us define local changes to $\gamma^{(k)}_j$ stemming from the genes by
\begin{align}
  \Delta_{ij}^{s,(k)} = G^{s}(\Phi(X_p),\bm{\gamma}^{(k)},\ldots,\bm{\gamma}^{(0)}), \label{eqn:Deltagene}
\end{align}
where $i$ is the contribution of the $i$th data set and $s$ denotes the chosen gene strategy~(see below for examples). Note that in the case of an empty data element $a_{ij}$,  one usually sets $\Delta_{ij}^{g}=0$, i.e. an empty data feature has no direct effect on the gene fitness. However, since all updates in Eq.\eqref{eqn:Deltagene} are relative, blank spaces can have an implicit effect on the dynamics of the evolutionary system.

Analogously, let us define the local changes to $\gamma^{(k)}_j$ stemming from the organisms by
\begin{align}
  \Delta_{ij}^{\omega,(k)} = \Omega^{s}(\Phi(X_p),\bm{\gamma}^{(k)},\ldots,\bm{\gamma}^{(0)}), \label{eqn:Deltaorganism}
\end{align}
where $i$ is the contribution of the $i$th data set and $s$ denotes the chosen organism strategy~(see below for examples). Note that one can interpret $\Delta_{ij}^{g}, \Delta_{ij}^{\omega}$ as relative surplus or deficiency of resources in the evolutionary simulation.

Hence, in general, the iterative updates $\bm{\gamma}^{(k)}\rightarrow \bm{\gamma}^{(k+1)}$ depend on the input data, the fitness function $\Phi$, the strategies and the history of gene fitness values $\bm{\gamma}^{(0)},\ldots,\bm{\gamma}^{(k)}$. In this paper, we will restrict ourselves to a simpler case, where only the previous gene fitness values are taken into account
\begin{align}
  \Delta_{ij}^{g,(k)} = G^{GS}(\Phi(X_p),\bm{\gamma}^{(k)}), \label{eqn:Deltagenesimple}\\
  \Delta_{ij}^{\omega,(k)} = \Omega^{OS}(\Phi(X_p),\bm{\gamma}^{(k)}). \label{eqn:Deltasorganismsimple}
\end{align}
As a next step we accumulate the contributions from all organisms for a certain gene
\begin{align}
  \Delta_{j}^{(k)} = \sum_{i=1}^n \left(\Delta_{ij}^{g,(k)}+\Delta_{ij}^{\omega,(k)}\right). \label{eqn:Deltageneaccumulated}
\end{align}
For convenience, let us assume that the strategies GS and OS are chosen that $\Delta_{j}^{(k)}$ is not changing to rapidly in each iteration
\begin{align}
  -1\leq\Delta_{j}^{(k)} \leq 1 \text{ for all genes } 1\le j \le m.
\end{align}

Let us now consider the general replicator equations for Genetic AI
\begin{align}
  \widetilde{\gamma}_j^{(k+1)} = \gamma_j^{(k)} \left(1+\Delta_{\gamma_j}^{(k)} \right),\label{eqn:replicator1}\\
  \gamma_j^{(k+1)} = \frac{\widetilde{\gamma}_j^{(k+1)}}{\sum_\ell^m \widetilde{\gamma}_\ell^{(k+1)}},\label{eqn:replicator2}
\end{align}
where the latter equation normalizes the gene fitness values after the updates.

Quite similar to EGT, the gene fitness values $\bm{\gamma}^{(k)}$ may converge to a evolutionary stable equilibrium~(ESE)\footnote{One usually wants a non-trivial ESE, where not all gene fitness is transferred to a single gene.}. Whether a non-trivial ESE is reached depends mainly on the evolutionary strategies and the population $\Phi(X_p)$, i.e. the data at hand. Even in non-converging and trivial cases, it might still pay to investigate the dynamics after a few iterations, because the speed of changes $\dot{\bm{\gamma}}^{(k)} = \partial\bm{\gamma}^{(k)}\/\partial k$ usually leads to understanding of the underlying data.

Another important boundary condition are the initial values $\bm{\gamma}^{(0)}$, which can be chosen either uniformly distributed or asymmetrical, taken into account preliminary knowledge about the system. Also in this case, it is convenient to stop the simulation \emph{before} a possible ESE is reached - to prevent that this preliminary knowledge is lost.

\section{Evolutionary Strategies}
\label{sec:evolutionarystrategies}

In Genetic AI, where the gene$\leftrightarrow$data-feature analogy acts as a framework for the data model and the replicator equations guarantee normalization, the evolutionary strategies mainly determine the dynamics of the system. Hence, a major part in the (further) development of Genetic AI boils down to analyzing and comparing strategies.

In general, genes and organisms compete with each other 'in terms' of their strategies. In this sense, organisms act as 'extended phenotype' wrt. their genes~\cite{Dawkins2006}. This is an important difference to purely statistical approaches in data analysis, since it may give individual data sets highly different importance in the simulation. Governed by the replicator equations Eq.~\eqref{eqn:replicator1}+\eqref{eqn:replicator2} there is a flow of fitness between the genes and organisms.

Though this 'game' of resources and fitness and the corresponding dynamics mimics the analogous behavior in EGT, there are also major differences: for once, all genes are competing against all other genes in every game round. Additionally, the behavior of a gene is determined by its underlying gene variant fitness and not by which 'opponent' it encounters. Consequently, the genes and organisms compete rather independently for a general heap of resources according to a global strategy. In the implementation, this algorithmic trait allows Genetic AI to be parallelized very easily.

Most evolutionary strategies are similar in what they do: apply a certain local data analysis function to a gene or organism, respectively. This function measures or compares properties of the input data $\Phi(X_p)$ from the 'perspective' of that gene or organism, respectively.

\subsection{Gene Strategy: Dominant}

As a first strategy for gene behavior, we define the GS-Dominant as
\begin{align}
\Delta_{ij}^{g:dom} = \frac{4\gamma_j^2}{n}\left[\phi_j(a_{ij})-\frac{1}{2}\right]. \label{eqn:gs-dominance}
\end{align}
Note that for all gene variant fitness values $\phi_j(a_{ij})$ bigger than 50\%,
the strategy will increase the gene fitness, in other cases it will lower the gene fitness~(see also Alg.~\ref{alg:GSdom} for a pseudo code of that strategy). Hence, genes with more gene variant fitness will dominate others who
have less.

From an evolutionary perspective the 'Dominant' gene strategy Eq.~\eqref{eqn:gs-dominance} describes the genes ability to reproduce itself~(hence the quadratic factor of gene fitness $\gamma_j^2$~\footnote{Note that the quadratic dependence to the gene fitness has fascinating analogies to formulas of electric charge interaction and methods to describe charge densities.}). The normalization factor $4$ stems is required for a fair comparison of $\Delta_{ij}^{g:dom}$ to organism strategies and other gene strategies. Also for the other strategies we will add a factor $2$ for every gene or organism fitness value included.

\begin{algorithm}[H]
\caption{GS-Dominant}
\begin{algorithmic}[1]
\Function{Update}{$\Phi(X_p)$, $\bm{\gamma}$, gene, GS} 
\ForAll{gene variants} 
\If{gene variant $> 50$\%}
\State increase gene fitness
\Else 
\State reduce gene fitness
\EndIf
\EndFor
\EndFunction
\end{algorithmic}
\label{alg:GSdom}
\end{algorithm}

\subsection{Organism Strategy: Balanced}

Since GS-Dominant depends in a sense on asymmetry, we require an organisms strategy that 'counteracts' with a balancing effect in order to allow for an ESE. Hence, we define OS-Balanced as
\begin{align}
\Delta_{ij}^{\omega:bal} = -\frac{2 r_i}{n}\left[\mu_{ij}-\frac{1}{m}\right],\label{eqn:os-balance}
\end{align}
where $\mu_{ij}$ is the contribution of the $j$th gene variant to the overall organism fitness
\begin{align}
  \mu_{ij} = \frac{\gamma_j \phi_j(a_{ij})}{r_i}.
\end{align}
Note that the expression in the brackets determines the sign
in Eq.~\eqref{eqn:os-balance}. If a particular gene variant contributes more than the $m$th part to an organisms' fitness, the value $\Delta_{ij}^{\omega:bal}$ becomes negative, reducing the genes fitness. Hence, an organism 'wants' to avoid being too dependent on a single gene in terms of its own fitness~(see also Alg.~\ref{alg:OSbal} for a pseudo code of that strategy). In the data picture this means that the relevance of data features dominating data sets gets a penalty and vice versa.

\begin{algorithm}[H]
\caption{OS-Balanced}
\begin{algorithmic}[1]
\Function{Update}{$\Phi(X_p)$, $\bm{\gamma}$, $\bm{\omega}$, OS} 
\ForAll{genes} 
\State $\mu\gets$ contribution of gene to organism fitness
\If{$\mu$ $> 1/$(number of genes)}
\State decrease gene fitness
\Else 
\State increase gene fitness
\EndIf
\EndFor
\EndFunction
\end{algorithmic}
\label{alg:OSbal}
\end{algorithm}

\subsection{Gene Strategy: Altruistic}

Let us now introduce a more complex strategy for genes: GS-Altruistic. The idea is that genes exchange fitness, depending on their 'kinship', i.e. their genetic similarity, and their fitness. To that end, let us define the~(symmetric) gene kinship between gene $j$ and~$\ell$ as
\begin{align}
 \label{eqn:genekinship}
 \kappa_{j\ell}^g  = 1-\frac{\norm{\bm{g}_j-\bm{g}_\ell}}{n}.
\end{align}
We now define GS-Altruistic as
\begin{align}
\tilde{\Delta}_{ij}^{g:alt} &= \frac{4}{m}\sum_{\ell, \ell\neq j}^m \gamma_\ell\kappa_{j\ell}^g \left[\phi_{\ell}(a_{i\ell})-\phi_{j}(a_{ij})\right], \label{eqn:gs-altruismtilde}\\
\Delta_{ij}^{g:alt} &= \frac{\Delta_{ij}^{g:dom}\cdot\tilde{\Delta}_{ij}^{g:alt}}{\gamma_j}, \label{eqn:gs-altruism}
\end{align}
where we use $\Delta_{ij}^{g:dom}$ from Eq.~\eqref{eqn:gs-dominance}. To understand the qualitative dynamics of this strategy it pays to investigate the signs of the contributions
\begin{align}
 \mathcal{G}^{alt}_{ij}=\left(\sign  \Delta_{ij}^{g:dom}, \sign \tilde{\Delta}_{ij}^{g:alt}\right).\label{eqn:genealtsigns}
\end{align}
If both contributions have the same sign, $\Delta_{ij}^{g:alt}$ becomes positive, leading to increased gene fitness. In contrast, opposite signs of  $\Delta_{ij}^{g:dom}, \tilde{\Delta}_{ij}^{g:alt}$ will decrease gene fitness. Omitting the cases where one factor is trivial, there are four scenarios:

\begin{description}
 \item $\mathcal{G}^{alt}_{ij}=(+,+)$: a dominant gene variant $\phi(a_{ij})$ has even more dominant relatives. These thus related genes will altruistically give fitness to the $j$th gene.
 \item $\mathcal{G}^{alt}_{ij}=(+,-)$: a dominant gene variant $\phi(a_{ij})$ has weaker relatives. These thus related genes will take fitness from the $j$th gene.
 \item $\mathcal{G}^{alt}_{ij}=(-,+)$: a recessive gene variant $\phi(a_{ij})$ has more dominant relatives. These thus related genes will take fitness from the $j$th gene.
 \item $\mathcal{G}^{alt}_{ij}=(-,-)$: a recessive gene variant $\phi(a_{ij})$ has even weaker relatives. These thus related genes will altruistically give fitness to the $j$th gene.
\end{description}


\subsection{Organism Strategy: Selfish}

Inversely to GS-Altruistic, we can also introduce the 'counter-acting' strategy OS-Selfish for organisms. To that end, we require the~(symmetric) organism kinship between gene~$i$ and~$t$
\begin{align}
 \label{eqn:orgkinship}
 \kappa_{it}^\omega  = 1-\frac{\norm{\bm{\omega}_i-\bm{\omega}_t}}{m}
\end{align}
and the initial range of organism fitness
\begin{align}
\label{eqn:irangeorgfitness}
 \rho = \max_{1\le t\le n} r_t^{(0)} - \min_{1\le t\le n} r_t^{(0)}
\end{align}
where we assume for simplicity that not all initial organism fitness values are equal~\footnote{In the case of equal organism fitness values, an alternative choice is $\rho=\max_{1\le t\le n} r_t^{(0)}$.}.

Then, OS-Selfish is defined as
\begin{align}
\tilde{\Delta}_{ij}^{\omega:sel} &= \frac{1}{n}\sum_{t, t\neq i}^m \kappa_{it}^\omega \frac{r_i-r_t}{\rho}, \label{eqn:os-selfishtilde}\\
\Delta_{ij}^{\omega:sel} &= \frac{\Delta_{ij}^{\omega:bal}\cdot\tilde{\Delta}_{ij}^{\omega:sel}}{2r_i}, \label{eqn:os-selfish}
\end{align}
where $r_t, r_i$ are the organism fitness values. Note that the initial range of organism fitness $\rho$ from Eq.~\eqref{eqn:irangeorgfitness} measures how close a call the evolutionary game is for the organisms. If $\rho$ is small, Eq.\eqref{eqn:os-selfishtilde} will yield larger transfers of resources between closely related organisms. In the evolutionary picture, the closer the organism are in fitness in a habitat, the more extreme they will apply their evolutionary advantages and strategies to preveil in the environment.

To understand the qualitative dynamics of this strategy let us again investigate the signs of the contributions
\begin{align}
 \mathcal{O}^{sel}_{ij}=\left(\sign  \Delta_{ij}^{\omega:bal}, \sign \tilde{\Delta}_{ij}^{\omega:sel}\right).\label{eqn:orgselsigns}
\end{align}

As for GS-Altruism, there are four scenarios:
\begin{description}
 \item $\mathcal{O}^{sel}_{ij}=(+,+)$: an undervalued gene $j$ is hosted by an organism that is dominating its relatives. The organism $i$ selfishly takes fitness from related organisms to increase the gene fitness $\gamma_j$.
 \item $\mathcal{O}^{sel}_{ij}=(+,-)$: an undervalued gene $j$ is hosted by an organism that is inferior to its relatives. The organism $i$ has to give fitness to related organisms and decreases the gene fitness $\gamma_j$.
 \item $\mathcal{O}^{sel}_{ij}=(-,+)$: an overvalued gene $j$ is hosted by an organism that is dominating its relatives. The organism $i$ decreases the gene fitness $\gamma_j$ to become more balanced.
 \item $\mathcal{O}^{sel}_{ij}=(-,-)$: an overvalued gene $j$ is hosted by an organism that is inferior to its relatives. For a stronger relative $t$, the contribution of the $j$th gene $\mu_{tj}$ tends to be less important, i.e. $\mu_{tj}<\mu_{ij}$. Therefore, the weaker organism $i$ selfishly takes fitness from its relatives and increases $\gamma_j$ to get fitter wrt. to its relatives.
\end{description}

\subsection{Choice of Strategies}

With a selection of gene and organism strategies at hand, the question arises: which pair of strategies should one choose? There are essentially three routes to proceed which we introduce in the next sections.

One major observation, which will become more apparent in the numerical experiments Sec.~\ref{sec:numericalexperiments}, is that the combination GS-Dominant+OS-Balanced mainly tests symmetries of the system, while GS-Altruistic+OS-Selfish mainly tests similarities or correlations of data. Hence, for a general data analysis taking into account both realms, we define
\begin{align} 
 \Delta_{ij}^g &= \alpha^{g}_{dom} \Delta_{ij}^{g:dom} + \alpha^{g}_{alt}\Delta_{ij}^{g:alt},\label{eqn:combiningdelta1}\\
 \Delta_{ij}^\omega &= \alpha^{\omega}_{bal} \Delta_{ij}^{\omega:bal} + \alpha^{\omega}_{sel} \Delta_{ij}^{\omega:sel},\label{eqn:combiningdelta2}\\
 \text{where } &\sum_s\alpha_s^g =\sum_s \alpha^\omega_s = 1.
\end{align}
Note that one may use a linear combination of an arbitrary number of strategies. Let us now come to three different ways to choose the coefficients $\alpha$.

\subsubsection{\emph{Ab initio} approach}
\label{sec:abinitioapproach}
Taking a closer look at the linear combinations~\eqref{eqn:combiningdelta1}+\eqref{eqn:combiningdelta2} it becomes apparent that the quantities $\Delta_{ij}^*$ have structural similarities with basis functions in solid state physics, e.g. in Kohn-Sham equations~\cite{Kohn1965}. While self-consistency in physical systems seems quite different to our evolutionary picture, it appears natural to look for a way to determine $\alpha$~(and, possibly, additional coefficients for further 'basis' strategies) from first principles. 

The difficulty of a fully self-consistent approach is that allowing a free competition between evolutionary strategies might jeopardize reaching an ESE at some time. In Sec.~\ref{sec:self-consistentdynamics}, we present one way to allow the gene fitness to converge for simple examples. In future work, with more strategies~('basis functions'), a more generic way for self-consistency might be beneficial.


\subsubsection{Predefined choice}
In real-world applications, one often understands the dominating behavior of the system the data describes. Hence, a predefined mixing of $\alpha$ or other strategies is a convenient choice. After leaving the strict \emph{ab initio} rules, we can also customize the initial gene fitness $\bm{\gamma}^{(0)}$ to match individual preferences. This provides a very easy method to take into account user preferences. It is important to not let the so disturbed system relax to a uniform ESE, but stop the simulation at an earlier stage.

\subsubsection{Determine the mixing through training}\label{sec:mixingtroughtraining}
Leaving the \emph{ab initio} approach completely, we can also use training data for a problem to determine the optimal mixing $\alpha$. In this case it pays to use the training data to determine the optimal combination of evolutionary strategies, but customize $\bm{\gamma}^{(0)}$ according to e.g. user preferences.

\section{Numerical Experiments}
\label{sec:numericalexperiments}

Before getting to the numerical tests, let us introduce some necessary examples for gene variant fitness functions. Apart from the boolean function Eq.~\eqref{eqn:booleangenevariant}, we will require the percentage fitness function
\begin{align}
  \phi_j^{per}(a_{ij}) = \frac{a_{ij}}{\displaystyle\max_{1\leq \ell \le 1} a_{\ell j}},
\end{align}
the inverse percentage fitness function
\begin{align}
  \phi_j^{inv}(a_{ij}) = 1-\frac{a_{ij}}{\displaystyle\max_{1\leq \ell \le 1} a_{\ell j}}.
\end{align}
Though we will only employ numeric gene variant fitness functions here, let us quickly give an example for labelled data
\begin{align}
  a_{ij} = [\text{label-1}, \text{label-2},\ldots, \text{label-t}].
\end{align}
Then, an example for an overlap fitness function could be
\begin{align}
  \phi_j^{over}[\text{label-2}, \text{label-t}](a_{ij}) = \frac{\Theta[\text{label-2}](a_{ij}) + \Theta[\text{label-t}](a_{ij})}{2},
\end{align}
where 
\begin{align}
\Theta[\text{label}](a_{ij})=\left\{
    \begin{array}{ll}
        0 & \text{if } a_{ij} \text{ does not contain the label,}\\
        1 & \text{if } a_{ij} \text{ contains the label.}\\
    \end{array}
    \right. \label{eqn:theta}
\end{align}

\subsection{Simple Example}
\begin{table}[tb]
  \caption{Simple example - input data $X$ for a small choice of flights.}
  \begin{tabular}{rccc}
    \toprule
          &price[Euro]&time-of-transfer[h]&stops\\
    \midrule
    flight A & $300$ &  $10$ & $2$\\
    flight B & $600$ &  $5$  & $2$\\
    flight C & $1500$ & $4$  & $1$\\
  \bottomrule
\end{tabular}
\label{tab:flight1}
\end{table}

As a first introductory example let us investigate a decision problem: choosing the right flight out of a list of $n=3$ offers with $m=3$ data features~(see Tab.\ref{tab:flight1} for the input data). Since for all data features in this example, larger is worse, we apply $\phi_j^{inv}$ for all columns to obtain the gene variant values of the population~(see Tab.\ref{tab:flight2}).

\begin{table}
  \caption{Simple example - population $\Phi(X_p)$ with 3 organisms $\bm{\omega}_A,\bm{\omega}_B,\bm{\omega}_C$ for a small choice of flights.}
  \begin{tabular}{rccc}
    \toprule
          &price[Euro]&time-of-transfer[h]&stops\\
    \midrule
    $\bm{\omega}_A$ & $0.8$ &  $0$ & $0$\\
    $\bm{\omega}_B$ & $0.6$ &  $0.5$  & $0$\\
    $\bm{\omega}_C$ & $0$ & $0.6$  & $0.5$\\
  \bottomrule
\end{tabular}
\label{tab:flight2}
\end{table}

In the most simplest case, the initial values for the gene fitness are symmetric
\begin{align}
  \bm{\gamma}^{(0)} = \left[\frac{1}{3}, \frac{1}{3}, \frac{1}{3}\right],
\end{align}
i.e. we have no initial preference in terms of data features. With Eq.~\eqref{eqn:linorganismfitness} this leads to the initial organism fitness of
\begin{align}
 \bm{r}^{(0)} = [0.266, 0.366, 0.366],
\end{align}
i.e. it is indecisive, whether flight B or C show a superior fitness.

Applying Eq.~\eqref{eqn:gs-dominance}, we obtain the gene resources matrix
\begin{align}
\Delta^{g:dom,(0)} = 
 \begin{pmatrix}
   \phantom{-}0.044      &       -0.074      &    -0.074     \\    
   \phantom{-}0.015      &       \phantom{-}0.000      &    -0.074 \\        
  -0.074      &       \phantom{-}0.015      &    \phantom{-}0.000  \\
 \end{pmatrix},
\end{align}
where we collect the contributions of all rows
\begin{align}
\label{eqn:simpleexampledeltag2}
\Delta^{g:dom,(0)}_j = \sum_{i=1}^n \Delta_{ij}^{g:dom,(0)}=[-0.01,-0.05,-0.15],
\end{align}
i.e. all genes are recessive in terms of the gene strategy GS-Dominant for this example.
Analogously, with Eq.~\eqref{eqn:os-balance}, we obtain the organism resources matrix
\begin{align}
\Delta^{\omega:bal,(0)} = 
 \begin{pmatrix}
   -0.119      &       \phantom{-}0.059      &  \phantom{-}0.059       \\    
   -0.052      &      -0.030      &  \phantom{-}0.082 \\        
   \phantom{-}0.082      &       -0.052      &  -0.030  \\
 \end{pmatrix},
\end{align}
where we again collect the contributions of all rows
\begin{align}
\label{eqn:simpleexampledeltao2}
\Delta^{\omega:bal,(0)}_j = \sum_{i=1}^n \Delta_{ij}^{\omega:bal,(0)}=[-0.09,-0.02,0.11].
\end{align}
With Eq.~\eqref{eqn:Deltageneaccumulated} we arrive at
\begin{align}
 \Delta_j^{(0)} = [-0.10, -0.08, -0.04].
\end{align}
One reason for all gene updates to be recessive, lies in the 'center of gravity' of the population, i.e. the average of all entries of Tab.~\ref{tab:flight2} being $0.33$ and thus, below $0.5$. In this sense, the $X$ from Tab.~\ref{tab:flight1} is a 'weak' population of data.
However, in Genetic AI, only relative values matter. Hence, we apply the replicator equations Eq.\eqref{eqn:replicator1}+\eqref{eqn:replicator2} to obtain
\begin{align}
 \bm{\gamma}^{(1)} = [0.32, 0.33, 0.35].
\end{align}
\begin{figure}[tb]
  \centering
  \includegraphics[width=\linewidth]{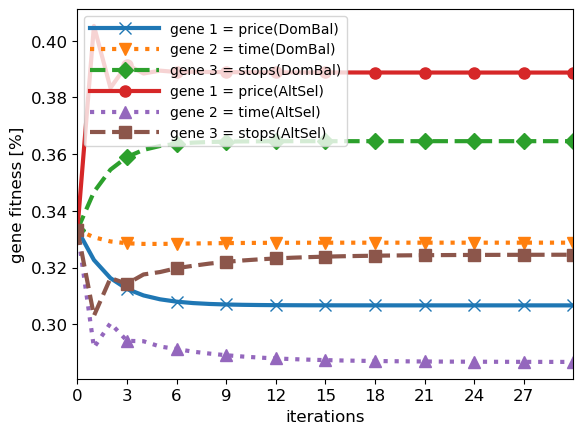}
  \caption{Gene fitness for data Tab.~\ref{tab:flight1} using GS-Dominant+OS-Balanced~(DomBal) or GS-Altruistic+OS-Selfish~(AltSel) after 30 iterations of evolutionary simulation.}
  \label{fig:3x3genefitness}
\end{figure}
\begin{figure}[tb]
  \centering
  \includegraphics[width=\linewidth]{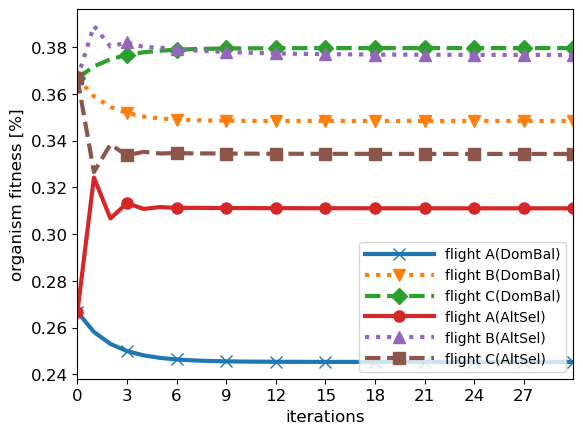}
  \caption{Organism fitness for data Tab.~\ref{tab:flight1} using GS-Dominant+OS-Balanced~(DomBal) or GS-Altruistic+OS-Selfish~(AltSel) after 30 iterations of evolutionary simulation.}
  \label{fig:3x3orgsfitness}
\end{figure}
Note that gene fitness is flowing from gene 'price' to gene 'stops' whereas gene 'time-of-transfer' remains more or less unchanged. Consequently, with Eq.\eqref{eqn:linorganismfitness}, this leads us to new values for the organism fitness
\begin{align}
 \bm{r}^{(1)} = [0.26, 0.36, 0.37],
\end{align}
i.e. flight C 'wins' the game. In Fig.\ref{fig:3x3genefitness}+\ref{fig:3x3orgsfitness}, one can observe the dynamics of the gene fitness and organism fitness after 30 iterations, respectively~\footnote{Note that the results for $i=0$ in Fig.~\ref{fig:3x3orgsfitness}, represents an example for a weighted approach as e.g. in some variants EMO\cite{Friedrich2011}, but with equal weights for all objectives. In contrast to EMO, where optimal solutions are investigated, we focus on the evolutionary dynamics of the weights, the gene fitness values $\bm{\gamma}$.}. The reason why the gene describing 'stops' is increasing in relevance, is that two-thirds of the data sets, the organisms $1+2$, have $0$ in this data feature~(see Tab.~\ref{tab:flight2}) and do not 'like' their dependence on the other genes. As can be seen in Eq.\eqref{eqn:simpleexampledeltag2}, though the strategy GS Dominant 'punishes' the gene 'stops', the organism strategy OS Balanced Eq.~\eqref{eqn:simpleexampledeltao2} ensures that this negative effect is more than remedied. In contrast, for the gene for 'price', both GS Dominant and OS Balanced return negative contributions - hence, 'price' turns out to be the gene of lowest fitness. Hence, this explains why 'stops' comes out dominant in this simple simulation.

%

Let us now apply GS-Altruistic+OS-Selfish to Tab.~\ref{tab:flight2}. In Fig.~\ref{fig:3x3genefitness}, we find the gene fitness values for $30$ iterations, again starting from a symmetric initial gene fitness $\gamma_1^{(0)}=\gamma_2^{(0)}=\gamma_3^{(0)}=1/3$. After the ESE is reached, the gene for 'price' dominates the population with $39$\%, followed by the gene 'stops'~($32$\%) and 'time'~($28$\%) comes out last.

To understand these results, let us investigate the matrix of contributions plugging the data from Tab.~\ref{tab:flight2} into Eq.~\eqref{eqn:gs-altruismtilde}-\eqref{eqn:genealtsigns}
\begin{align}
\mathcal{G}^{alt} = 
 \begin{pmatrix}
   (+,-)      &       (-,+)      &    (-,+) \\    
   (+,-)      &       (0,+)      &    (-,+) \\        
   (-,+)      &       (+,-)      &    (0,+) \\
 \end{pmatrix},
\end{align}
i.e. all contributions $\Delta_{ij}^{g:alt}$ are negative. Turning to the organism contributions from Eq.~\eqref{eqn:os-selfishtilde}-\eqref{eqn:orgselsigns}
\begin{align}\label{eqn:simpleexamplecontributionmatrix}
\mathcal{O}^{sel} = 
 \begin{pmatrix}
   (-,---)      &       (+,--)      &    (+,--) \\    
   (-,+)      &       (-,+)      &    (+,+) \\        
   (+,+)      &       (-,+)      &    (-,+) \\
 \end{pmatrix},
\end{align}
where multiple signs denote larger contributions. Here, we see that the first gene/column 'price' gets two positive contributions $i=1,i=3$, the second gene/column 'time' has only decreases, while the third gene/column 'stops' has one increasing contribution for $i=2$. 

Accumulating, the gene 'price' gets one very large positive contribution, one small positive and four small negative contributions. The gene 'time' gets four~(smaller) negative contributions and one larger negative term. And, finally, the gene 'stops' gets one large negative contribution, three smaller negative and one smaller positive contributions. Hence, the order price$>$stops$>$time of the strategies AltSel in Fig.~\ref{fig:3x3genefitness} can be understood.

In addition to the order of gene fitness, application of AltSel to the simple example~\ref{tab:flight2} also shows another important observation: how correlation comes into play for these strategies. In particular, the gene kinship
\begin{align}
\kappa^{g} = 
 \begin{pmatrix}
   1      &       0.67      &    0.63 \\    
   0.67      &       1      &    0.83 \\        
   0.63      &      0.83      &    1 \\
 \end{pmatrix}
\end{align}
shows that the genes 'time'<>'stops' are highly correlated with $\kappa_{12}^{g}=0.83$. Why do these two genes behave so differently? For this dynamics, the second flight $\omega_B$ takes a crucial role. As can be seen in Tab.\ref{tab:flight2}, flight B has a gene variant value of $0.5$ for 'time' and $0$ for 'stops'. Thus, $\mathcal{O}^{sel}_{23}=(+,+)$ for 'stops' stems from a highly undervalued gene rests in an organism dominating its relatives, whereas $\mathcal{O}^{sel}_{22}=(-,+)$ follows from $0.5$ for 'time' being overvalued for $\omega_B$. In summary, whereas flight A wins the game for 'price', flight B decides the second place for 'stops'.

In conclusion, AltSel appears to test for correlations in the input data. Though this observation seems trivial in this small example, similar behavior can also be seen in cases with much larger input data. In particular, Genetic AI provides a way to measure multi-dimensional, cascading data correlations through evolutionary simulation.


\subsection{Real-World Example}
\label{sec:realworldexample}

\begin{figure}[tbp]
  \centering
  \includegraphics[width=\linewidth]{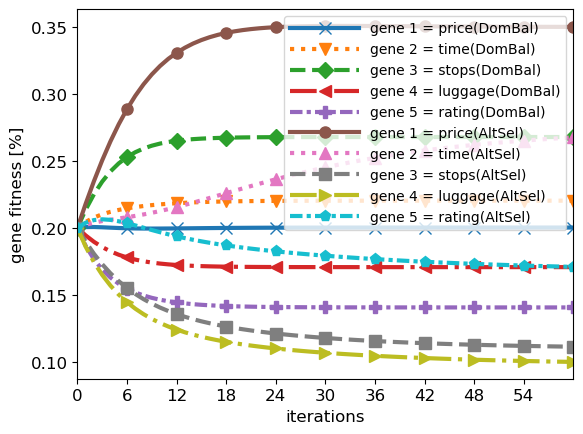}
  \caption{Gene fitness for the real-world example Tab.~\ref{tab:flight3} using GS-Dominant+OS-Balanced~(DomBal) or GS-Altruistic+OS-Selfish~(AltSel) after 500 iterations of evolutionary simulation.}
  \label{fig:10x5genefitness}
\end{figure}

\begin{figure}[tbp]
  \centering
  \includegraphics[width=\linewidth]{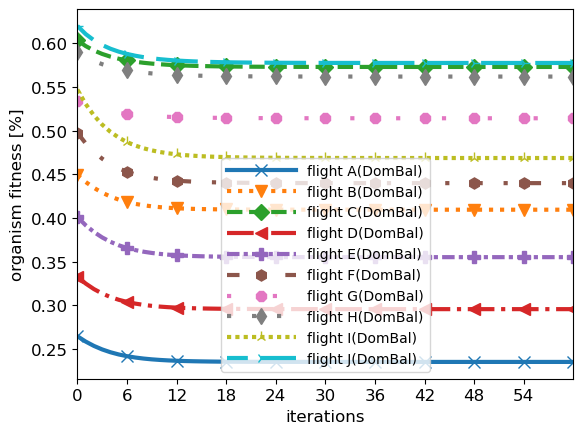}
  \caption{Organism fitness for the real-world example Tab.~\ref{tab:flight3} using GS-Dominant+OS-Balanced~(DomBal) after 65 iterations of evolutionary simulation.}
  \label{fig:GSdomOSbal10x5orgfitness}
\end{figure}


\begin{figure}[tb]
  \centering
  \includegraphics[width=\linewidth]{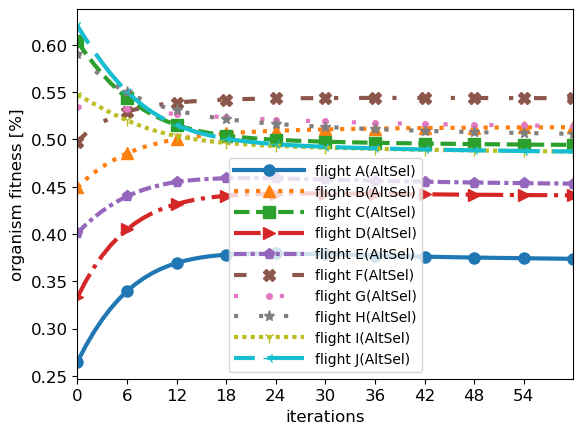}
  \caption{Organism fitness for the real-world example Tab.~\ref{tab:flight3} using GS-Altruistic+OS-Selfish~(AltSel) after 500 iterations of evolutionary simulation.}
  \label{fig:GSaltOSsel10x5orgfitness}
\end{figure}

\begin{table}[tbh]
  \caption{Real-world example - input data $X$ for 10 flights and 5 data features}
  \label{tab:flight3}
   \begin{tabular}{lccccc}
     \toprule
            &price[Euro]&time[h]&stops&luggages&rating\\
      \midrule
       flight A &  $300$ & $10$ & $2$ & $0$ & $2.5$\\
       flight B &  $600$ &  $5$ & $2$ & $1$ & $3.0$\\
       flight C & $1500$ &  $4$ & $1$ & $2$ & $4.0$\\
       flight D &  $400$ &  $8$ & $2$ & $0$ & $3.5$\\
       flight E &  $500$ &  $8$ & $2$ & $1$ & $3.0$\\
       flight F &  $700$ &  $5$ & $2$ & $1$ & $4.5$\\
       flight G &  $900$ &  $6$ & $1$ & $1$ & $4.0$\\
       flight H & $1100$ & $6$ & $1$& $2$& $3.5$\\
       flight I & $1300$ & $5$ & $2$& $2$& $5.0$\\
       flight J & $1700$ & $4$ & $1$& $2$& $5.0$ \\
    \bottomrule
 \end{tabular}
\end{table}


Let us now populate our example Tab.\ref{tab:flight1} with additional data features and flight options. First, we introduce the number of luggage a passenger can take with him/her. Second, we take the airline rating of customer satisfaction. Additionally, we add $7$ additional flight options and arrive at the data shown in Tab.~\ref{tab:flight3}.

In Fig.~\ref{fig:10x5genefitness}, we show the evolution of the gene fitness for the five genes in this example using GS-Dominant+OS-Balanced. In this case, in the ESE, the gene 'stops' outperforms all others with $26$\% fitness. This can be understood since six out of ten flights have $2$ stops, which leads to minimal gene variant fitness $\phi_3(a_{i3})=0$. For this majority of 6 flights, the organism strategy OS Balanced yields positive values, since the these flights want to remedy their minimal dependence on this gene. Summarizing, 'stops' becomes the most important gene since many organisms with larger fitness interpret two stops as their main flaw - hence it gets upvoted.

The gene minimal with the minimal fitness in this example turns out to be 'rating'. This is because of an opposite effect wrt. 'stops'. Since most flights have good rating, they get downvoted in relevance by OS Balanced. Since most flights with good fitness have a good rating anyway, it seems not so relevant for a decision using the strategies 'DomBal'.

In Fig.~\ref{fig:GSdomOSbal10x5orgfitness} we show the evolution of organism fitness for the $10$ flights from Tab.~\ref{tab:flight3} using GS-Dominant+OS-Balanced. Though all flights lose fitness, the flight J dominates the population with a fitness of more than $57$\%. Though flight J is the most expensive at $1700$ Euro, this is the only property it does not have the maximum value. Since the gene 'price' gets an intermediate fitness shown with 'DomBal' in Fig.~\ref{fig:10x5genefitness}, the thus obtained fitness penalty is not enough the counter the excellent values of flight J in the other data features.

Now we want to apply AltSel to the data in Tab.~\ref{tab:flight3}. As can be seen in Fig.~\ref{fig:10x5genefitness}, the gene price hugely outperforms all other genes in this case. For the dynamics in this example, the behavior of a single organism, flight A, plays a crucial role. In the beginning, since its fitness is by far the weakest, flight A selfishly shifts a massive amount of fitness. In the first iteration, its contribution to the fitness updates compared to the total updates is
\begin{align}
 \Delta^{\omega:sel}_{1j} &= [0.040 -0.019, -0.019, -0.019, 0.017],\\
 \Delta^{\omega:sel} &= [0.091, -0.023, -0.030, -0.055, 0.016],
\end{align}
i.e. flight A determines the sign and amounts to approximately $40-90\%$ of the total updates to gene fitness from organisms. Calculating the contribution signs from flight A
\begin{align}
 \mathcal{O}^{sel}_{1j} = 
 \begin{pmatrix}
   (--,-)      &       (+,-)      &    (+,-) & (+,-) & (-,-)
 \end{pmatrix}
\end{align}
where '$--$' means again a larger contribution. Generally, it can be seen that, as by far the weakest organism, flight A, selfishly shifts large chunks of fitness from the genes $2-4$ to gene $1$, where the balanced and the selfish part of Eq.~\eqref{eqn:os-selfish} work in the same direction. With this flow of gene fitness it can be seen in Fig.~\ref{fig:GSaltOSsel10x5orgfitness} that the organism flight A becomes fitter at the expense of other organisms during the simulation. In the end, the flight F benefits from this dynamics most and wins this evolutionary game.

\subsection{Fully self-consistent evolutionary dynamics}
\label{sec:self-consistentdynamics}

In the example above, we predefined the evolutionary strategies and investigated the dynamics of the gene and organism fitness. What happens when we let the data system itself choose the appropriate strategies? In this section, we want to do this last step to a complete \emph{ab initio} approach and demonstrate the effects on our two introduced examples.

Let us first define which set of gene strategies and organism strategies we include in our self-consistent cycle
\begin{align}
 \mathcal{S}_g &= \{ dom, alt \},\\
 \mathcal{S}_\omega &= \{ bal, sel \}.
\end{align}
These strategies yield individual delta contributions according to Eq.~(\ref{eqn:gs-dominance}+\ref{eqn:gs-altruism}+\ref{eqn:os-balance}+\ref{eqn:os-selfish}). The principle idea behind our self-consistent approach is that we measure the absolute fitness effect of a strategy with
\begin{align}
\bar{\Delta}_s^g&=m\cdot\underset{1\le i,j \le n,m}{\mean}  |\Delta_{ij}^{g:s}| \text{ for } s\in\mathcal{S}_g,\label{eqn:scmixingdeltagene}\\
 \bar{\Delta}_s^\omega&=\underset{1\le i \le n}{\mean} \sum_{j=1}^m |\Delta_{ij}^{\omega:s}|\gamma_j \text{ for } s\in\mathcal{S}_\omega.\label{eqn:scmixingdeltaorg}
\end{align}

To iteratively determine the mixing factors $\alpha$ from Eq.(\ref{eqn:combiningdelta1}+\ref{eqn:combiningdelta2}) we now use the replicator equations analogous to Eq.(\ref{eqn:replicator1}+\ref{eqn:replicator2})
\begin{align}
  \widetilde{\alpha}_s^{(k+1)} = \alpha_s^{(k)} \left(1+\bar{\Delta}_{s}^{(k)} \right),\label{eqn:scmixing1}\\
  \alpha_s^{(k+1)} = \frac{\widetilde{\alpha}_s^{(k+1)}}{\sum_{t\in \mathcal{S}} \widetilde{\alpha}_t^{(k+1)}},\label{eqn:scmixing2}
\end{align}
which can be applied for $\bar{\Delta}_{s}^{g,(k)}, \bar{\Delta}_{s}^{\omega(k)}$.

\begin{figure}[tb]
  \centering
  \includegraphics[width=\linewidth]{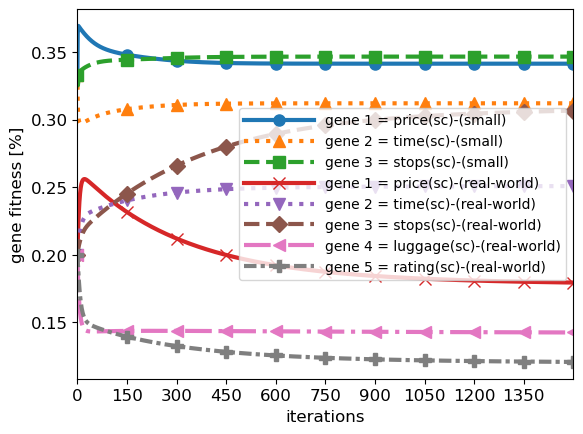}
  \caption{Fully self-consistent simulation comparing the gene fitness of the simple example Tab.~\ref{tab:flight1} and the real-world example Tab.~\ref{tab:flight3} .}
  \label{fig:scgenefitness}
\end{figure}

\begin{figure}[tb]
  \centering
  \includegraphics[width=\linewidth]{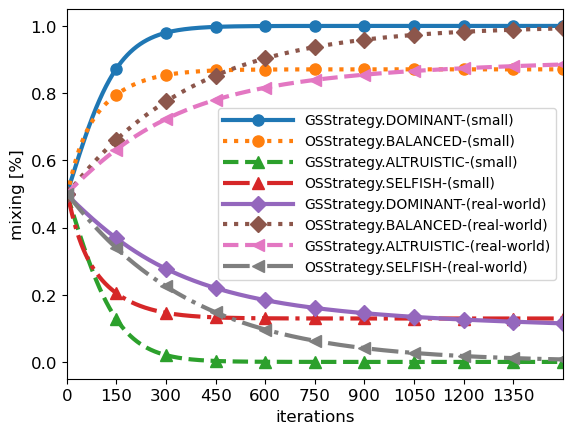}
  \caption{Fully self-consistent simulation showing the dynamics of the mixing factors~(compare also to Fig.\ref{fig:scgenefitness}).}
  \label{fig:scmixing}
\end{figure}

In Fig.~\ref{fig:scgenefitness}, we show the dynamics of the gene fitness for  fully self-consistent simulations of the small example and the real-world example, where we adapt the $\alpha$ values dynamically, according to Eq.(\ref{eqn:scmixingdeltagene}-\ref{eqn:scmixing2})). As a starting value, we set $\alpha_s^{(0)}=0.5$ for all strategies. Not surprisingly, convergence takes longer as in the cases with fixed strategies as both the mixing factors $\alpha$ \emph{and} the gene fitness values $\gamma$ have to settle to an equilibrium. 

Let us compare the values of $\bar{\Delta}_s^g$ after one iteration for the small example from Tab.~\ref{tab:flight1}
\begin{align}
 \bar{\Delta}_{dom}^{g,(1)} & = 0.041,\\
 \bar{\Delta}_{alt}^{g,(1)} & = 0.037
\end{align}
where it can be seen that GS Dominant has a stronger average effect than GS Altruistic and in the end represses the alternative strategy completely as can be seen in Fig.~\ref{fig:scmixing}.

Investigating $\bar{\Delta}_s^\omega$ for the two strategies
\begin{align}
 \bar{\Delta}_{bal}^{\omega,(1)} & = 0.062,\\
 \bar{\Delta}_{sel}^{\omega,(1)} & = 0.071,\\
 \bar{\Delta}_{bal}^{\omega,(2)} & = 0.065,\\
 \bar{\Delta}_{sel}^{\omega,(2)} & = 0.053,
\end{align}
we see that the initial tendency to OS Selfish is turned around for the second iteration. This behavior is caused by the change in the contribution of flight C
\begin{align}
\Delta^{\omega,(1)}_{sel,3j} &= [0.0640, -0.0407, -0.0233], \\
\Delta^{\omega,(2)}_{sel,3j} &= [0.0143, -0.0086, -0.0057].
\end{align}
Recalling the part of Eq.~\eqref{eqn:simpleexamplecontributionmatrix} for flight C
\begin{align}
\mathcal{O}^{sel}_{3j} = 
 \begin{pmatrix}
   (+,+)      &       (-,+)      &    (-,+) \\
 \end{pmatrix},
\end{align}
it follows that since flight C loses fitness wrt. both flights A and C, respectively, the most expensive flight also loses much potential to effect the overall gene fitness values dramatically. In particular flight B, which is at equal fitness in the first iteration, starts dominating flight C and starts selfishly reducing its contributions.

After 500 iterations, the GS Dominant has completely replaced GS Altruistic resulting in the gene for 'price' losing ground to 'stops'. This has the side effect that the contributions of flight C to OS Selfish can partly recover
\begin{align}
\Delta^{\omega,(500)}_{sel, 3j} &= [0.040, -0.022, -0.017]
\end{align}
as the competitive advantage of the flight A and B in terms of a lower price slowly loses its effect. As ESE is reached there is a mixed equilibrium of 87\% OS Balanced and 13\% OS Selfish as can be seen in Fig.~\ref{fig:scmixing}.

For the real-world example the results are different: whereas this time the strategy OS Balanced completely takes over with 100\% dominance, the strategies GS Altruistic and GS Dominant stabilize at a ratio of 89\%/11\%, respectively.

Why does GS Altruistic become the predominant strategy for the real-world example~Tab.~\ref{tab:flight3}, but not for the simple example Tab.~\ref{tab:flight1}? Recalling Eq.~\eqref{eqn:gs-altruism}, that, since GS Altruistic contains GS Dominant, it will yield larger contributions if $|\tilde\Delta^{g:alt}_{ij}|/\gamma_j>1$. Only two contributions fulfill this condition for the small example with the maximum $|\tilde\Delta^{g:alt}_{23}|/\gamma_3=1.14$ from Eq.~\eqref{eqn:gs-altruismtilde}
\begin{align}
 \frac{4}{m\gamma_3}\left\{\kappa_{21}^g\gamma_1[\phi_1(a_{21})-\phi_3(a_{23}]+ \kappa_{23}^g\gamma_2[\phi_2(a_{22})-\phi_3(a_{23}]\right\}.
\end{align}
Hence, the maximum value for GS Altruistic is only 14\% larger than GS Dominant. The main reason for this is that the genes relatives for flight B are not really strong with $\phi_1(a_{21})=0.6$ and $\phi_1(a_{22})=0.5$.

The case is different for the real-world example~\ref{tab:flight3}: here, the maximum factor for GS Altruistic is $|\tilde\Delta^{g:alt}_{10,1}|/\gamma_1=2.05$, i.e. twice as large as GS Dominant. This stems mainly from the two additional gene relatives 'luggage' and 'stops' which have the maximum value for flight J and therefore max-out the term for the transfer of resources $\phi_2(a_{10,4})-\phi_3(a_{10,1}$.

\subsection{Heuristic Observations}

Let us summarize the conclusions from the presented numerical experiments. To obtain the optimal solution and measure the relevance of data features, evolutionary strategies test various data properties. Every strategy acts like a test kernel comparing the principle data features of a given data package:

\begin{itemize}
 \item GS Dominant mainly tests for symmetry and individual gene variant fitness. It follows from Eq.~\eqref{eqn:gs-dominance} that if the majority of gene fitness values are above or below $0.5$, it will have the strongest effect on the dynamics of the simulation. On its own, GS Dominant would accumulate all gene fitness at the in this sense strongest gene.
 \item GS Altruistic Eq.~\eqref{eqn:gs-altruism} measures how similar or related different genes are. It will prevail over GS Dominant if there are enough related genes that have multiple statistical outliers which behave against a given similarity between data features.
 \item OS Balanced Eq.~\eqref{eqn:os-balance} measures the symmetry of individual data sets. If the fitness of a data set is distributed very unevenly, it will yield larger contributions. Thus, OS Balanced also acts against fitness accumulation coming from the gene strategies. It is thus a major reason for stable states that allow for the non-trivial data analysis.
 \item OS Selfish Eq.~\eqref{eqn:os-selfish} mainly tests how similar or related different organisms or data sets are. Note that if data sets tend to be rather similar, also their fitness values will be close to each other. Hence, OS Selfish can act as a tiebreaker in simulations that are otherwise to close to call.
 \item With this set of four evolutionary strategies, there was a non-trivial ESE in all examples that were investigated. In particular, it seems plausible that OS Balanced~(and the corresponding term in OS Selfish) will prevent a complete accumulation of the entire gene fitness to one single gene.
\end{itemize}

\subsection{Other Applications}

In the introduced decision examples, we investigated problems in terms of the dynamics of gene fitness and organism fitness, respectively. The same approach can be easily generalized for a large set of other fields. For example, Genetic AI can be used in search engines, recommendation and prediction. We leave the analysis and comparison to existing methods for future work however.

\section{Conclusion}
In this paper, we have introduced Genetic AI, a new method for optimization and data analysis from first principles. Applying Genetic AI to two simple decision problems, we have shown that it is a versatile tool to understand a system described by data. 

In the end, a question persists: why does it work? To understand this, it helps to completely turn to the evolutionary picture: assume we have a closed evolutionary system of genes and organisms with fixed preconditions. Lets further assume that there are no gene mutations, cross-overs or other changes to the genes and organisms of the population. Then, the competition of the genes and organisms of the system turns into a 'game' of how good the given properties perform in the chosen environment. On the one hand, this predefined environment 'tests' our fixed genes and organisms~(in Genetic AI, the strategies GS+OS mimic this environment to 'test' the data). On the other hand, the performance of genes and organisms is governed by universal mechanics of gene/organism correlations, similarities and symmetries. Note that only some of these mechanics are based on statistical dynamics, as single genes or organisms might change the outcome of the evolutionary game completely.

Quite analogously to evolutionary systems, data problems depend on correlations, similarities and symmetries between one part of the data to the others. Hence, we can expect evolutionary simulations to describe data systems, if we include all necessary evolutionary strategies to cover their fundamental behavior. 

\section{Outlook}
In ML and EA, it took around 30 years after their first introduction for wide-spread acceptance and applications. Hence, in the next years, we want to expand on the original formalism described in this paper.

One major objective is a detailed comparison of Genetic AI with different algorithms from the field of EMO~\cite{Friedrich2011}. Furthermore, a conceptual analysis of the relationship to EGT~\cite{MaynardSmith1973} would provide additional insights on the potential of Genetic AI to analyze universal data models. Creating hybrid systems, e.g. in connection to general Neural Networks or LLMs, might provide a powerful strategy to use the strengths of both technologies.

In terms of applications, it would be very useful to investigate larger, more complex and also more diverse examples to better pinpoint weaknesses and advantages of the method. To that end, we aim to create a 'network' of interconnected evolutionary simulations. Quite similar to organs in a human body, these individual simulations could independently investigate certain prerequisite questions and 'channel' their analysis into one final simulation, the 'brain'.


\vspace*{0.5cm}
\section*{Acknowledgments}
Thanks to Richard Allmendinger and Karsten Held for useful discussions and input. Thanks also Aris Daniilidis and his group for helpful discussions and inspirations. Also thanks to Martin Bär for proofreading, suggestions and references.

\bibliography{geneticai_v1}

\onecolumngrid

\end{document}